\documentclass[11pt]{article}

\usepackage[final]{acl}

\usepackage{times}
\usepackage{latexsym}
\usepackage[T1]{fontenc}
\usepackage[utf8]{inputenc}
\DeclareUnicodeCharacter{2192}{\ensuremath{\rightarrow}}
\usepackage{microtype}
\IfFileExists{inconsolata.sty}{\usepackage{inconsolata}}{}
\usepackage{graphicx}
\usepackage{booktabs}
\usepackage{amsmath}
\usepackage{amssymb}
\usepackage{comment}
\usepackage{tikz}
\usetikzlibrary{positioning, arrows.meta, shapes.geometric, calc, fit}
\usepackage{makecell}
\usepackage[table]{xcolor}

\newcommand{\yes}{\textcolor{green!50!black}{\ensuremath{\boldsymbol{\checkmark}}}}
\newcommand{\no}{\textcolor{red!70!black}{\ensuremath{\boldsymbol{\times}}}}
\newcommand{\partly}{\textcolor{orange!85!black}{\ensuremath{\boldsymbol{\sim}}}}

\title{Quizzing the Translation: A Prover-Grounded Evaluation Metric for NL→FOL}

\author{Pu Suo \\
  Emory University \\
  \texttt{psuo2@emory.edu} \And
  Ali Emami \\
  Emory University \\
  \texttt{aemami@emory.edu}}

\begin{document}
\maketitle
\begin{abstract}
A standard pipeline for symbolic reasoning over natural-language problems translates them into first-order logic and invokes a theorem prover. The translation step is the bottleneck: swap ``every'' for ``some'' and every inference that follows is corrupted. Yet today's metrics often score more broken translations higher than less broken ones, because BLEU, BERTScore, and Smatch++ reward surface overlap that the worst errors happen to preserve. We introduce SIV, which derives two kinds of probes from the target formula and uses a theorem prover to verify the candidate translation against each. \emph{Positive probes} are statements the candidate must entail, which detect translations that drop content; \emph{contrastive probes} are statements the candidate must \emph{not} entail, which detect translations that assert more than the original. On a controlled pool of perturbed FOLIO translations, the severity of the error accounts for $80\%$ of SIV's score variance, compared with at most $17\%$ for any prior metric. Across six error classes on a disjoint pool, SIV scores the reference above the perturbed candidate in over $99\%$ of pairs. Because each probe is labeled with what it tests, the failure pattern also supplies a labeled error trace, recovering the perturbation class at macro-F1 $0.638$, nearly double the score-only baseline. On $434$ expert-audited real LLM translations, SIV attains the top AUC, uniquely detects \emph{and} grades expert-labeled major errors, and abstains, rather than mis-scoring, on out-of-vocabulary translations.
\end{abstract}

\begin{figure}[!t]
\centering
\begin{tikzpicture}[
    font=\scriptsize,
    perturb/.style={draw=blue!50!black, fill=blue!4, rounded corners=2pt,
                    inner sep=4pt, align=left},
    metricpanel/.style={draw=gray!50!black, fill=gray!6, rounded corners=2pt,
                        inner sep=4pt, align=left},
    sivpanel/.style={draw=blue!55!black, fill=blue!10, rounded corners=2pt,
                     inner sep=4pt, align=left},
    panelwidth/.style={text width=68mm},
]
\node[perturb, panelwidth] (top) {%
\textbf{NL.} ``Holding companies hold several companies.''
\par\smallskip
$\varphi\;\;\:=\; \forall x.\,HC(x) \rightarrow \exists y.(C(y) \wedge H(x,y))$
\par\vspace{1pt}
{\scriptsize\hspace*{3em}$\downarrow$\; inner quantifier strengthened ($\exists\,{\to}\,\forall$)}
\par\vspace{1pt}
$\varphi' =\; \forall x.\,HC(x) \rightarrow \textcolor{red!75!black}{\boldsymbol{\forall}}\, y.(C(y) \rightarrow H(x,y))$%
};
\node[metricpanel, panelwidth, below=1.4mm of top] (mid) {%
\textbf{What existing metrics return:}
\par\smallskip
\setlength{\tabcolsep}{4pt}
\begin{tabular}{@{}lrl@{}}
BLEU       & $0.76$ & \no \\
BERTScore  & $0.87$ & \no \\
Smatch++   & $0.94$ & \no \ {\scriptsize(rates near-perfect)} \\
LE         & $0.75$ & \no \ {\scriptsize(no localization)} \\
LT         & $0.00$ & \yes \ {\scriptsize(binary; no localization)} \\
\end{tabular}%
};
\node[sivpanel, panelwidth, below=1.4mm of mid] (bot) {%
\textbf{What SIV does.} Derive probes from $\varphi$; adjudicate each with the prover.
\par\smallskip
{\itshape Positives} (candidate must entail):
\par\vspace{1pt}
\setlength{\tabcolsep}{3pt}
\begin{tabular}{@{}r@{\;}l@{\quad}l@{}}
$\varphi' \models$ & $\forall x.(HC(x) \rightarrow \exists y.\,C(y))$              & \yes \\
$\varphi' \models$ & $\forall x.(HC(x) \rightarrow \exists y.(C(y) \wedge H(x,y)))$ & \yes \\
\end{tabular}
\par\smallskip
{\itshape Contrastives} (candidate must not entail):
\par\vspace{1pt}
\setlength{\tabcolsep}{3pt}
\begin{tabular}{@{}r@{\;}l@{\quad}l@{}}
$\varphi' \models$     & $\forall x.(HC(x) \rightarrow \forall y.(C(y) \rightarrow H(x,y)))$ & \no \\
$\varphi' \not\models$ & $\forall x.\,\exists y.(C(y) \wedge H(x,y))$                         & \yes \\
$\varphi' \not\models$ & $\forall x.(HC(x) \rightarrow \neg\exists y.(C(y) \wedge H(x,y)))$  & \yes \\
$\varphi' \models$     & $\exists y.(C(y) \wedge \forall x.(HC(x) \rightarrow H(x,y)))$       & \no \\
\end{tabular}
\par\smallskip
\textbf{SIV-F1 $= 0.667$}\quad{\scriptsize (recall $2/2$, precision $2/4$)}
\par\vspace{1pt}
{\scriptsize trace: two contrastives fired, \texttt{flip\_quantifier} (inner $\exists\,{\to}\,\forall$) and \texttt{scope\_swap} ($\forall\exists\,{\to}\,\exists\forall$)}
};
\end{tikzpicture}
\caption{An inner-quantifier-strengthened FOLIO translation: ``every holding company holds every company'' rather than ``some.'' SIV is the only metric to both flag the error and name its structural axes.}
\label{fig:worked-example}
\end{figure}
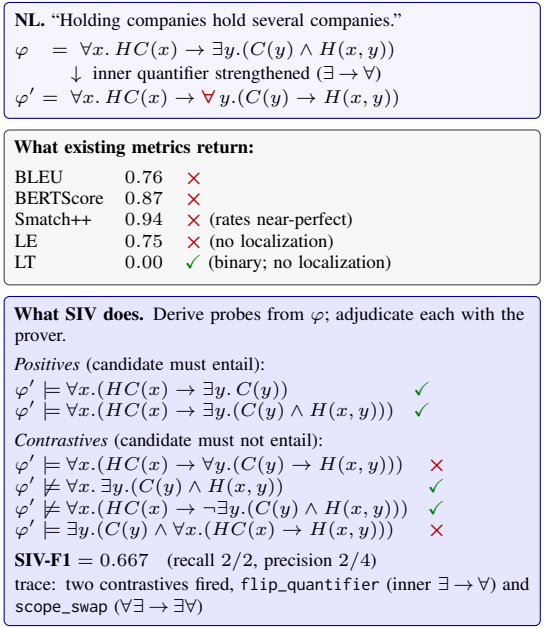

\section{Introduction}
\label{sec:intro}


Translating natural language into first-order logic is the bottleneck of any system that combines natural language with symbolic inference, and the most visible recent context is the LLM-prover pipeline pursued by Logic-LM \citep{pan-etal-2023-logiclm} and its successors \citep{olausson-etal-2023-linc, ye-etal-2023-satlm, ryu-etal-2025-clover}.  A single misplaced quantifier suffices to corrupt every inference that follows. Consider the FOLIO \citep{han-etal-2024-folio} premise ``Holding companies hold several companies,'' whose expert formalization is $\varphi = \forall x.\,HC(x) \rightarrow \exists y.(C(y) \wedge H(x,y))$. Strengthening the inner existential to a universal converts ``holds some company'' into ``holds every company'': downstream reasoning is now warranted by a claim the original never made.

Detecting such errors is the job of a translation metric, and the metrics currently used for NL$\rightarrow$FOL fall short in a characteristic way. Surface-level metrics such as BLEU \citep{papineni-etal-2002-bleu} and BERTScore \citep{zhang-etal-2020-bertscore} compare formulas as token strings or embeddings, and the graph-based Smatch++ \citep{opitz-2023-smatchpp} compares them as graphs. The Logical Equivalence (LE) score of \citet{yang-etal-2024-mallspaper} compares propositional truth tables, and the logical-theory (LT) check of \citet{brunello-2026-llms} returns a binary FOL-equivalence verdict via Z3. Figure~\ref{fig:worked-example} shows what each returns on the over-strengthened translation above: most rate it nearly correct (Smatch++ at $0.94$). The deeper problem, documented in \S\ref{sec:exp1}, is that these failures are not random. On a controlled pool of perturbed translations, BLEU, BERTScore, and Smatch++ systematically score severely-broken translations \emph{higher} than partially-broken ones, because the perturbations that damage meaning the most also tend to preserve the most surface content. More wrong is not more surface-different.

We propose \emph{Structural Inference Verification} (SIV), a metric that evaluates a translation by quizzing it. SIV checks whether the candidate is logically equivalent to its reference: whether it asserts neither more nor less than the reference does. A candidate can fail this in exactly two ways: by \emph{understating} the reference (asserting less; the reference entails the candidate but not vice versa) or by \emph{overstating} it (asserting more; the candidate entails the reference but not vice versa).

SIV catches both by deriving two kinds of probes from the reference and adjudicating each with the theorem prover Vampire \citep{kovacs-voronkov-2013-vampire}. \textbf{Positive probes} are statements the reference itself entails; the candidate must entail them too, or it has understated. \textbf{Contrastive probes} are statements stronger than the reference; the candidate must \emph{not} entail them, or it has overstated. SIV's score is the F1 over both probe sets, and each failed probe is labeled with the structural feature it tested, yielding a per-translation diagnostic trace. The Method section (\S\ref{sec:metric}) gives worked examples of each probe type.

We establish four findings:
\begin{itemize}
\setlength{\itemsep}{2pt}
\item \textbf{Existing metrics do not track error severity.} On a stratified FOLIO pool, BLEU, BERTScore, and Smatch++ assign higher scores to severely-broken translations than to partially-broken ones, and LE collapses to chance (AUC $= 0.500$) on quantifier flips. No prior NL$\rightarrow$FOL metric tracks how broken a translation actually is.
\item \textbf{SIV's score tracks it precisely.} Error severity accounts for $80\%$ of SIV's score variance, compared with at most $17\%$ for any baseline. Across six structural error classes on a disjoint pool of $1{,}865$ verified perturbation pairs, SIV ranks the reference above the perturbed candidate in over $99\%$ of cases. This is the architectural payoff of the two-sided design: on strictly-stronger errors, positive probes are saturated by construction and the contrastive arm carries the entire detection signal.
\item \textbf{SIV's trace labels the structural error.} Each failed probe is labeled with the structural feature it tests, so the pattern of failures classifies the error type, a signal useful for debugging translators, ranking failures, or supplying class-conditional feedback in training loops. A deterministic, parameter-free rule classifier over the trace recovers the class at six-class macro-F1 $= 0.638$, nearly twice the $0.333$ of the strongest score-only baseline.
\item \textbf{SIV transfers to real translator outputs.} On $434$ expert-audited LLM translations (\S\ref{sec:exp4}), SIV attains the top AUC point estimate ($0.841$), is the only metric that both detects and grades expert-labeled major errors ($0.20$ vs.\ $0.80$ on correct), and is the only metric with an out-of-vocabulary abstention flag (F1 $0.87$).
\end{itemize}

We release the parser, the probe suites, the expert-audited real-translation pairs, and full reproduction code.\footnote{Code and data: \url{https://github.com/pu-suo/siv-metric}.}

\begin{table}[t]
\centering
\setlength{\tabcolsep}{2pt}
\renewcommand{\arraystretch}{1.15}
\footnotesize
\begin{tabular}{l*{5}{c}}
\toprule
& \rotatebox[origin=l]{55}{\textbf{Prover-grnd.}}
& \rotatebox[origin=l]{55}{\textbf{Quant.-aware}}
& \rotatebox[origin=l]{55}{\textbf{Graded}}
& \rotatebox[origin=l]{55}{\textbf{Per-trans.}}
& \rotatebox[origin=l]{55}{\textbf{Diagn.\ trace}} \\
\midrule
BLEU      & \no & \no & \yes & \yes & \no \\
BERTScore & \no & \no & \yes & \yes & \no \\
Smatch++  & \no & \no & \yes & \yes & \no \\
LE        & \partly & \no & \yes & \yes & \no \\
LT        & \yes & \yes & \no & \yes & \no \\
EPR       & \yes & \yes & \no & \no & \no \\
\midrule
\rowcolor{blue!8}\textbf{SIV (ours)} & \yes & \yes & \yes & \yes & \yes \\
\bottomrule
\end{tabular}
\caption{Capability comparison of NL$\rightarrow$FOL evaluation metrics. \yes\ supported; \no\ not supported; \partly\ partially supported. SIV is the only metric combining all five.}
\label{tab:comparison}
\end{table}

\section{Related Work}
\label{sec:related}

\subsection{NL$\rightarrow$FOL translation and benchmarks}

MALLS \citep{yang-etal-2024-mallspaper} and FOLIO \citep{han-etal-2024-folio} anchor the NL$\rightarrow$FOL benchmark landscape. MALLS is GPT-4-generated at sentence scale and evaluated through the LE score on predicate truth tables; FOLIO is expert-written and evaluated through downstream entailment classification. \citet{thatikonda-etal-2026-improving} provide a taxonomy of translation errors and improve small-LM symbolic translation via error-aware fine-tuning. \citet{brunello-2026-llms} decompose NL$\rightarrow$FOL into ontology extraction and logical translation, showing that frontier dialogue-oriented LLMs perform the latter at over $0.90$ accuracy when given a fixed signature.

\subsection{Metrics for NL$\rightarrow$FOL translation}
\label{sec:related-metrics}

Table~\ref{tab:comparison} summarizes existing metrics along the capability axes for per-translation evaluation.

\paragraph{Existing metric families.} Surface metrics (BLEU, \citealp{papineni-etal-2002-bleu}; BERTScore, \citealp{zhang-etal-2020-bertscore}) and the graph-based Smatch \citep{cai-knight-2013-smatch} and Smatch++ \citep{opitz-2023-smatchpp} compare formulas at the token, embedding, or graph level, and cannot represent quantifier scope, binding, or polarity; \citet{funakura-etal-2025-tp-eval} and \citet{thatikonda-etal-2025-fol-metrics} both show empirically that graph-overlap does not predict logical equivalence. The LE score \citep{yang-etal-2024-mallspaper} reduces FOL to propositional truth tables, discarding quantifier structure. The LT check \citep{brunello-2026-llms} adjudicates first-order equivalence via Z3 correctly, but returns only a binary verdict: no gradient, and no localization of the error. EPR, the Entailment Preservation Rate of \citet{lee-etal-2025-epf}, is likewise prover-grounded but reference-free and corpus-level: it scores a translator by whether prover inference over the translated premise set reproduces gold entailment labels, and so provides neither a per-translation score nor a diagnostic.

\paragraph{Methodological precedent.} Test-suite evaluation for SQL semantic parsing \citep{zhong-etal-2020-semantic,wang-etal-2024-toolsql} is the closest precedent: derive a suite of behavioral queries from the reference and grade each. We port the idea from SQL to FOL by replacing query execution with prover-adjudicated entailment.

\paragraph{Adjacent metrics on different problems.} Tree-edit metrics with logical rewrites (GTED, \citealp{liu-etal-2025-gted}; ASSESS, \citealp{liu-etal-2025-assess}) target Lean autoformalization; edit distance cannot tell an over-claiming candidate from an equivalent one, the case SIV's contrastive probes handle by design. Predicate alignment \citep{hosseini-etal-2018-predicate} and Jaccard partial correctness \citep{wong-mooney-2007-semantic} evaluate vocabulary or clause-set overlap rather than prover-checked entailment.

\paragraph{Where SIV fits.} SIV combines capabilities that prior metrics distribute across families: prover-grounded adjudication (as in LT and EPR), a graded score (as in surface and graph metrics), and a per-translation diagnostic trace that no prior metric provides. All three come from a single mechanism: decomposing the reference into a labeled set of probes and adjudicating each independently. A principled \emph{graded} notion of logical similarity is difficult in general \citep{paris-vencovska-2015}; SIV makes a reference-anchored version tractable by reducing gradation to counts over independently adjudicated, prover-certified probes.

\label{sec:related-critiques}


\section{The SIV Metric}
\label{sec:metric}

\begin{figure}[t]
\centering
\begin{tikzpicture}[
    >={Stealth[length=1.3mm]},
    semithick,
    every node/.style={font=\scriptsize, align=center,
                       rounded corners=2pt, draw, inner sep=2.5pt},
    refbox/.style   ={fill=blue!10,   draw=blue!55!black,   text width=22mm, minimum height=5mm},
    candbox/.style  ={fill=blue!10,   draw=blue!55!black,   text width=16mm, minimum height=5mm},
    procbox/.style  ={fill=gray!12,   draw=gray!60!black,   text width=28mm, minimum height=7mm},
    bigprocbox/.style={fill=gray!12,  draw=gray!60!black,   text width=36mm, minimum height=8mm},
    posbox/.style   ={fill=green!13,  draw=green!55!black,  text width=24mm, minimum height=8mm},
    negbox/.style   ={fill=red!10,    draw=red!60!black,    text width=24mm, minimum height=8mm},
    outbox/.style   ={fill=orange!15, draw=orange!75!black, text width=24mm, minimum height=7mm},
]
\node[refbox] (ref) at (0, 0) {Reference $\varphi$};
\node[procbox] (parser) at (0, -0.85) {Parser};
\node[posbox] (pp) at (-1.6, -2.05) {Positive probes $\mathcal{P}^{+}$ \\ \tiny weakenings of $\varphi$};
\node[negbox] (pn) at ( 1.6, -2.05) {Contrastive probes $\mathcal{P}^{-}$ \\ \tiny perturbations of $\varphi$};
\node[bigprocbox] (vamp) at (0, -3.3) {Vampire prover \\ \tiny $\varphi' \models p$?};
\node[candbox] (cand) at (-3.0, -3.3) {Candidate $\varphi'$};
\node[outbox] (score) at (-1.4, -4.45) {SIV score \\ \tiny $F_1$};
\node[outbox] (trace) at ( 1.4, -4.45) {Trace \\ \tiny verdict + label};
\draw[->] (ref) -- (parser);
\draw[->] (parser) -- (pp);
\draw[->] (parser) -- (pn);
\draw[->] (pp) -- (vamp);
\draw[->] (pn) -- (vamp);
\draw[->] (cand) -- (vamp);
\draw[->] (vamp) -- (score);
\draw[->] (vamp) -- (trace);
\end{tikzpicture}
\caption{SIV pipeline. Vampire adjudicates the candidate $\varphi'$ against positive and contrastive probes derived from $\varphi$; the verdicts produce an F1 score and a labeled trace.}
\label{fig:overview}
\end{figure}
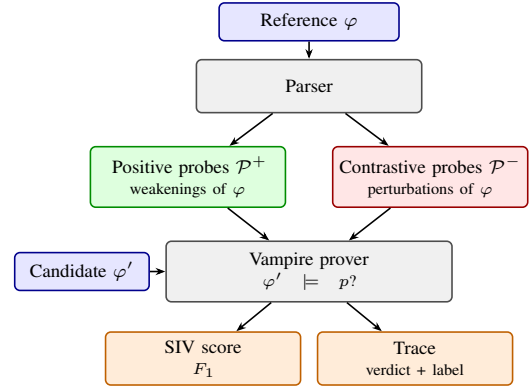

\subsection{Overview}
\label{sec:metric-setup}

A translation is correct when it is logically equivalent to its reference $\varphi$: it must entail what $\varphi$ entails (otherwise it \emph{understates}), and must not entail what $\varphi$ does not (otherwise it \emph{overstates}). SIV catches each failure mode with its own probe set; Figure~\ref{fig:overview} sketches the full pipeline. We use a running example throughout: the sentence \emph{``Every car has a black, round wheel''} with reference
\begin{equation}
\label{eq:phi-cars}
\begin{aligned}
\varphi = \forall x.\big( & \mathit{Car}(x) \rightarrow \exists y.(\mathit{Wheel}(y) \\
& \wedge\, \mathit{HasWheel}(x,y) \\
& \wedge\, \mathit{Black}(y) \wedge \mathit{Round}(y))\big).
\end{aligned}
\end{equation}
The inner existential $\exists y.(\mathit{Wheel}(y) \wedge \ldots)$ has typing predicate $\mathit{Wheel}(y)$ as its \emph{restrictor}, and the remaining body conjuncts $\mathit{HasWheel}(x,y), \mathit{Black}(y), \mathit{Round}(y)$ as its \emph{nuclei}; this is the standard (restrictor, nucleus, scope) decomposition of generalized quantifiers \citep{barwise-cooper-1981}.\footnote{Produced by a deterministic CFG parser; round-trip parsing yields a logically equivalent formula on $99.94\%$ of FOLIO premises (Vampire-verified).}

\paragraph{Positive probes ($\mathcal{P}^{+}$)} catch understatement. Each is a formula entailed by $\varphi$ that isolates one piece of its structure. For \eqref{eq:phi-cars}, $\mathcal{P}^{+}$ contains the reference itself together with four further probes: the restrictor-only ``every car has some wheel,'' plus three probes that keep wheel-existence and add one further conjunct each, with the outer universal-implication scaffolding preserved at every step:
\begin{align*}
& \forall x.(\mathit{Car}(x) \rightarrow \exists y.\,\mathit{Wheel}(y)),\\
& \forall x.(\mathit{Car}(x) \rightarrow \exists y.(\mathit{Wheel}(y) \wedge \mathit{Black}(y))),\\
& \forall x.(\mathit{Car}(x) \rightarrow \exists y.(\mathit{Wheel}(y) \wedge \mathit{Round}(y))),\\
& \forall x.(\mathit{Car}(x) \rightarrow \exists y.(\mathit{Wheel}(y)\\
& \qquad \wedge\, \mathit{HasWheel}(x,y))).
\end{align*}
A correct candidate must entail all five. A failure on any one is a specific weakness: missing the wheel-existence commitment, missing blackness, missing roundness, or missing the having-relation.

\paragraph{Contrastive probes ($\mathcal{P}^{-}$)} catch overstatement. Each is a perturbation of $\varphi$ that is either a strictly-stronger statement (entailing $\varphi$ but not entailed by $\varphi$) or one incompatible with $\varphi$ (jointly contradictory). A correct candidate must \emph{not} entail any. Dropping the outer restrictor $\mathit{Car}(x)$ from \eqref{eq:phi-cars} yields a strictly-stronger contrastive,
\begin{equation*}
\begin{aligned}
\forall x.\,\exists y.( & \mathit{Wheel}(y) \wedge \mathit{HasWheel}(x,y) \\
& \wedge \mathit{Black}(y) \wedge \mathit{Round}(y)),
\end{aligned}
\end{equation*}
which asserts that \emph{every} entity, not just every car, is associated with such a wheel. A candidate that entailed this would be making a structural commitment $\varphi$ does not.

How each probe set is built is the topic of \S\ref{sec:metric-positive} and \S\ref{sec:metric-contrastive}; \S\ref{sec:metric-scoring} defines the score. Every probe also carries a label naming the structural feature it tests, which feeds the per-translation diagnostic trace evaluated in \S\ref{sec:exp3}.

\paragraph{Scope.} SIV evaluates \emph{structural fidelity within an alignable logical vocabulary}, not arbitrary semantic equivalence. Candidate predicates are aligned to the reference vocabulary where possible (Appendix~\ref{app:alignment}); pairs whose vocabulary cannot be grounded are flagged, not silently scored (\S\ref{sec:exp4}).

\subsection{Building positive probes}
\label{sec:metric-positive}

Positive probes are generated by recursive descent through $\varphi$. At each binder, implication, or conjunction, we emit one probe per droppable conjunct, preserving the surrounding context; disjunctions are never decomposed. A disjunction only ever enters $\mathcal{P}^{+}$ carried whole, as a single dropped conjunct of an enclosing conjunction, which is a sound weakening; dropping a \emph{disjunct} strengthens a formula and therefore belongs to the contrastive inventory (\texttt{disjunct\_drop}, Appendix~\ref{app:operators}). Concretely, for an existential-conjunction $\exists y.(R(y) \wedge N_1(y) \wedge \cdots \wedge N_k(y))$ inside scope $S[\cdot]$, with restrictor $R$ and nuclei $N_i$, we emit the restrictor-only probe $S[\exists y.\,R(y)]$ and the restrictor-plus-one-nucleus probes $S[\exists y.(R(y) \wedge N_i(y))]$ for each $i$. Universal quantifications and implications decompose analogously.

Preserving the surrounding scope is what makes the construction sound: every emitted probe is a syntactic weakening of $\varphi$, so $\varphi \models p$ holds by construction for every $p \in \mathcal{P}^{+}$. In the cars example, the outer $\forall x.(\mathit{Car}(x) \rightarrow \cdot)$ is carried into every probe. Without it, the bare ``$\exists y.\,\mathit{Wheel}(y)$'' would not be entailed by $\varphi$ (a domain with no cars makes $\varphi$ vacuously true but says nothing about wheels), so this probe would not appear in $\mathcal{P}^{+}$. Every emitted probe is also Vampire-certified against its reference at suite-build time (zero-failure audit, Appendix~\ref{sec:probe-stats}), so an unsound probe cannot enter the suite.

\subsection{Building contrastive probes}
\label{sec:metric-contrastive}

Contrastive probes are generated in two steps: we apply structural operators to $\varphi$, then check each output against $\varphi$ before admitting it into $\mathcal{P}^{-}$. We use ten operators, including \texttt{swap\_binary\_args} (reversing relational argument order), \texttt{flip\_quantifier} (swapping $\forall \leftrightarrow \exists$ on a single binder), \texttt{drop\_restrictor\_conjunct} (the operator for the cars-example contrastive in \S\ref{sec:metric-setup}), \texttt{negate\_atom} (flipping the polarity of an atomic predicate), and six others spanning quantifier scope, disjunct removal, connective change, and equality (full inventory in Appendix~\ref{app:operators}). Each operator declares whether its output should be strictly-stronger than $\varphi$ or incompatible with it, and a type-soundness check discards ill-formed outputs. The inventory was selected for systematic coverage of FOL's structural axes rather than curated from errors observed during development; Appendix~\ref{app:operators} documents per-operator provenance, the selection principle, and the inventory's measured coverage boundary on real translator outputs (\S\ref{sec:exp4}).

Operators are only generators: an operator's declared relation to $\varphi$ is a prediction, not a guarantee. The theorem prover Vampire \citep{kovacs-voronkov-2013-vampire} checks the prediction before admission. A strictly-stronger $\psi$ is admitted only if $\psi \models \varphi$ and $\varphi \not\models \psi$; an incompatible $\psi$ only if $\varphi \wedge \psi \models \bot$. The check runs under Vampire's default axioms plus per-predicate and per-restrictor existence witnesses derived from $\varphi$; for binary relations, asymmetry and symmetry axioms come from a frozen $427$-predicate label table, so that for $R$ labeled asymmetric, $\forall x, y.\,R(x,y) \to \neg R(y,x)$ certifies \texttt{swap\_binary\_args} as incompatible on ground-atom occurrences. A perturbation that turns out to be logically independent of $\varphi$ (Vampire returns neither entailment nor contradiction) is dropped. The bias is conservative: we accept lost contrastive coverage rather than risk a probe whose declared relation does not hold. \S\ref{sec:exp3-stratum} returns to the model-theoretic limits of this regime.

\subsection{Scoring}
\label{sec:metric-scoring}

Let $T^+ = \{p \in \mathcal{P}^+ : \varphi' \models p\}$ be the positive probes the candidate correctly entails, and $F^- = \{p \in \mathcal{P}^- : \varphi' \not\models p\}$ the contrastive probes it correctly does not. Treating ``should be entailed'' as the positive class of a binary classifier, the SIV score is the corresponding F1:
\begin{equation}
\mathrm{SIV}(\varphi', \varphi) = \frac{2\,|T^+|}{|T^+| + |\mathcal{P}^+| + |\mathcal{P}^-| - |F^-|}.
\end{equation}
Concretely, a candidate that omits the blackness commitment entails three of the five positive probes from \S\ref{sec:metric-setup} and fails two (Black and the reference itself); being strictly weaker than $\varphi$, it entails no contrastive, so $|F^-| = |\mathcal{P}^-|$ and $\mathrm{SIV} = 6/8 = 0.75$, with the $0.25$ drop localized to the two failed probes, both naming Blackness.

Both probe arms are necessary. The positive arm alone treats a strictly-stronger candidate as equivalent to a correct one (both entail every positive probe), so the contrastive arm is what catches overstatement. The contrastive arm alone treats a vacuous candidate (one that entails nothing) as equivalent to a correct one (neither entails any contrastive), so the positive arm is what catches understatement.

We report the \emph{soft} variant of SIV, which treats Vampire timeouts symmetrically (a $5$-second per call timeout); a \emph{strict} variant that treats timeouts as ``not entailed'' produces identical drop magnitudes on \S\ref{sec:exp2}'s perturbation pool. The full FOLIO probe suite contains $2{,}340$ positive and $3{,}932$ contrastive probes over $1{,}393$ premises.

\paragraph{Cost.} Probe construction and Vampire admission are paid \emph{once per reference} ($\sim 0.35$\,s) and reused across every candidate and system; the marginal cost of scoring one additional candidate is $67$\,ms single-core, $6.8$\,ms on $14$ cores (zero timeouts in $14{,}293$ calls on the \S\ref{sec:exp2} pool). Evaluating $N$ systems thus costs build-once plus $N \times$ seconds; the released pipeline ships prebuilt suites. Appendix~\ref{app:runtime} profiles cheaper variants and one pitfall.

\section{Experiment 1: Severity tracking}
\label{sec:exp1}

This experiment tests whether SIV's score is calibrated to structural severity: candidates more distant from the reference should receive lower scores, and the magnitude of the drop should track the magnitude of the structural change.

\begin{table}[t]
\centering
\setlength{\tabcolsep}{3pt}
\renewcommand{\arraystretch}{1.15}
\footnotesize
\begin{tabular}{llll}
\toprule
\textbf{Tier} & $\varphi'$ \textbf{vs.\ }$\varphi$ & \textbf{Skeleton} & \textbf{Example perturbation} \\
\midrule
\textsc{Ref} & equivalent        & ---     & canonical re-serialization \\
\textsc{OS}  & strictly stronger & intact  & add a nucleus conjunct \\
\textsc{P}   & strictly weaker   & intact  & drop a nucleus conjunct \\
\textsc{OW}  & strictly weaker   & broken  & flip outer $\forall \to \exists$ \\
\bottomrule
\end{tabular}
\caption{Severity tiers, ordered by structural distance from $\varphi$. \emph{Skeleton} refers to the top-level quantifier-connective pattern. \textsc{OW} ranks below \textsc{P} because skeleton damage changes the formula's structural meaning, not just its informational content. \textsc{Ref} candidates are logically equivalent canonical re-serializations of the gold (Vampire-verified); their surface form differs from the FOLIO gold string (\S\ref{sec:exp1}, Setup).}
\label{tab:tiers}
\end{table} 
\subsection{Setup}

\paragraph{Tiers.}
We sort candidates into four severity tiers by their logical relation to the reference $\varphi$ (Table~\ref{tab:tiers}). \textsc{Ref} candidates are \emph{not} byte-identical copies of the gold: each is the reference's own AST re-serialized to canonical ASCII FOL (predicate names preserved). All $105$ are Vampire-certified equivalent to their gold in both directions, and $104/105$ differ textually from the gold string, which is why surface metrics score \textsc{Ref} below $1.0$ (Table~\ref{tab:correlation-and-means}). \textsc{OS} (overstrong) and \textsc{P} (partial) each represent one unit of structural change in opposite directions: \textsc{OS} preserves all of $\varphi$'s content but adds a strengthening commitment, whereas \textsc{P} drops content from $\varphi$ while preserving the top-level skeleton. \textsc{OW} (overweak) is worse than \textsc{P} because the same weakening change additionally breaks the skeleton, changing what the formula structurally means rather than merely what it asserts. Appendix~\ref{app:tier-construction} gives full tier construction rules.

\paragraph{Pool.}
We apply $14$ typed perturbation operators across five formula strata to FOLIO premises. Each operator declares its target tier; every generated candidate is bidirectionally verified by Vampire and retained only if its actual logical relation to the reference matches that declaration. The final pool contains $372$ candidates ($105$ \textsc{Ref}, $58$ \textsc{OS}, $25$ \textsc{P}, $184$ \textsc{OW}) drawn from $128$ FOLIO premises.\footnote{Train split. SIV is evaluated against synthetic perturbations rather than learned, so no test partition is held out.}

\paragraph{Statistics.}
$\eta^2$ from one-way ANOVA with severity tier as the factor, candidate-level on the three non-\textsc{Ref} tiers ($n=267$), measures the score variance attributable to severity; Cohen's $d$ on \textsc{Ref}-vs-\textsc{OW} measures the endpoint gap, aggregated to per-(premise, tier) means over the $53$-premise subset with $\geq 2$ non-\textsc{Ref} tiers, with $1{,}000$-resample bootstrap CIs. We use magnitude statistics rather than rank correlation because \textsc{OS}/\textsc{P} share rank by design.

\paragraph{Baselines.}
BLEU \citep{papineni-etal-2002-bleu} (tokenized at the formula level); BERTScore \citep{zhang-etal-2020-bertscore} (F1 on stringified formulas); Smatch++ \citep{opitz-2023-smatchpp} (graph-overlap F1 on a deterministic FOL$\to$Penman conversion with bound variables anonymized); LE \citep{yang-etal-2024-mallspaper} (truth-table comparison with greedy literal binding); LT \citep{brunello-2026-llms} (Z3 binary FOL equivalence). LT is excluded from $\eta^2$ because its binary range makes both statistics degenerate.

\subsection{Results}

\begin{table}[t]
\centering
\setlength{\tabcolsep}{4pt}
\renewcommand{\arraystretch}{1.15}
\footnotesize
\begin{tabular}{lcccccc}
\toprule
& \textbf{Ref} & \textbf{OS} & \textbf{P} & \textbf{OW} & $\eta^2$ & $d$ \\
\textbf{Metric} & \scriptsize{$105$} & \scriptsize{$58$} & \scriptsize{$25$} & \scriptsize{$184$} & & \\
\midrule
\rowcolor{blue!8}\textbf{SIV} & $\mathbf{1.00}$ & $\mathbf{.929}$ & $\mathbf{.442}$ & $\mathbf{.052}$ & $\mathbf{.802}$ & $\mathbf{4.71}$ \\
LE & $1.00$ & $.850$ & $.796$ & $.713$ & $.075$ & $1.50$ \\
LT (bin.) & $1.00$ & $.000$ & $.000$ & $.000$ & --- & --- \\
LT (graded) & $1.00$ & $.500$ & $.500$ & $.500$ & $.000^{\ddagger}$ & --- \\
Smatch++ & $1.00$ & $.806$ & $.629$ & $.740^{\dagger}$ & $.072$ & $1.84$ \\
BLEU & $.485$ & $.396$ & $.315$ & $.397^{\dagger}$ & $.027$ & $1.09$ \\
BERTScore & $.872$ & $.815$ & $.729$ & $.822^{\dagger}$ & $.170$ & $1.35$ \\
\bottomrule
\end{tabular}
\caption{Per-tier means and effect sizes. $\eta^2$: variance explained by tier. $d$: Cohen's $d$ on Ref-vs-OW. $\dagger$: metric scores OW \emph{higher} than P. LT binary; $\eta^2$ and $d$ are degenerate. $\ddagger$: graded-LT (bidirectional prover entailment scored $1/0.5/0$; \S\ref{sec:exp1-robust}) is constant across the three error tiers, so its between-tier variance is zero. Bootstrap $95\%$ CIs on $d$: SIV $[3.96, 6.04]$, LE $[1.29, 1.83]$, Smatch++ $[1.59, 2.19]$, BERTScore $[1.02, 1.71]$, BLEU $[0.72, 1.51]$; SIV's interval does not overlap any baseline's.}
\label{tab:correlation-and-means}
\end{table}

\textbf{Severity tier explains $80\%$ of SIV's score variance; at most $17\%$ for any baseline.} SIV reaches $\eta^2 = 0.80$; the strongest baseline (BERTScore) reaches $0.17$. SIV's \textsc{Ref}-vs-\textsc{OW} effect size is Cohen's $d = 4.71$, more than $2.5\times$ the strongest baseline (Smatch++, $1.84$), with non-overlapping CIs (Table~\ref{tab:correlation-and-means}).

\begin{figure}[t]
\centering
\includegraphics[width=\linewidth]{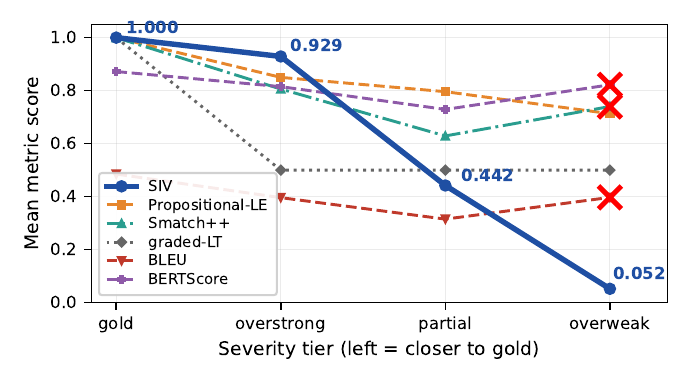}
\caption{Mean metric score by severity tier (Table~\ref{tab:correlation-and-means}). SIV descends monotonically (\textsc{Ref}$\to$\textsc{OW}: $1.00 \to 0.05$); Smatch++, BLEU, and BERTScore invert at \textsc{P}$\to$\textsc{OW} (red $\times$); graded-LT is constant at $0.5$ across all error tiers (\S\ref{sec:exp1-robust}). Binary LT omitted.}
\label{fig:severity-curve}
\end{figure}

\textbf{Surface and graph metrics invert at \textsc{P}$\to$\textsc{OW}.} BLEU, BERTScore, and Smatch++ each score the more-broken \textsc{OW} tier \emph{higher} than \textsc{P} (Table~\ref{tab:correlation-and-means}, $\dagger$; Figure~\ref{fig:severity-curve}). \textsc{P} weakens content inside an intact skeleton; \textsc{OW} damages the skeleton yet often preserves more surface tokens (a quantifier flip changes one symbol; a nucleus drop deletes a whole conjunct), so token overlap ranks \textsc{OW} above \textsc{P}, against severity. \emph{More wrong is not more surface-different.}

\textbf{LE collapses to chance on quantifier operators.} The per-operator AUC decomposition (Appendix~\ref{app:per-operator-auc}) localizes LE's $\eta^2 = 0.075$: on \texttt{flip\_outer\_quantifier} and \texttt{strengthen\_quantifier}, LE separates the perturbed tier from \textsc{Ref} at AUC $= 0.500$ (chance), while SIV reaches $1.000$. This reproduces \citet{brunello-2026-llms}'s propositional-collapse prediction operator by operator: LE's truth-table representation discards quantifier structure.

\textbf{BLEU charges half its range for re-serialization.} BLEU scores the reference's prover-verified logical twin at $0.485$; re-serializing both sides through the same printer restores it to $1.000$ on all $105$ premises. A metric that spends half its range on notation is not measuring logical content.

\textbf{The calibration is robust to design choices.}\phantomsection\label{sec:exp1-robust} Four checks (Appendix~\ref{app:robustness}): (i)~merging the equal-rank \textsc{OS}/\textsc{P} tiers leaves SIV at $\eta^2 = 0.706$ vs.\ at most $0.07$ for any baseline, with zero severity-order reversals across all $372$ candidates; (ii)~restricting to the pool operators structurally \emph{disjoint} from SIV's contrastive inventory leaves $\eta^2 = 0.759$, $d = 4.37$ $[3.76, 5.49]$, still ${\sim}5\times$ any baseline; (iii)~the reference's direct \emph{negation} receives SIV $0.022$, below the lowest tier, while Smatch++ scores the same complement $0.864$, above every error tier; (iv)~a \emph{graded LT} (bidirectional entailment under SIV's prover setup, scored $1/0.5/0$) detects every \S\ref{sec:exp2} class yet is constant at $0.5$ across all three error tiers (Table~\ref{tab:correlation-and-means}, $\ddagger$) and yields no trace: the prover supplies detection; gradation and diagnosis come from the two-sided probe decomposition.

\section{Experiment 2: Per-class detection}
\label{sec:exp2}

Experiment 1 measured aggregate calibration across error types. Experiment 2 disaggregates this by structural error class, testing whether SIV ranks the reference above the perturbed candidate and how large the score gap is.

\subsection{Setup}

\paragraph{Pool.}
We apply six typed perturbation operators to a $642$-premise FOLIO reference set, disjoint by construction from Experiment 1's $128$-premise design pool. After Vampire bidirectional verification, $1{,}865$ verified (reference, perturbed) pairs remain across six classes (Table~\ref{tab:exp2-detection}). The classes span the three logical relations a perturbed candidate can have to its reference: \emph{incompatible} (the candidate disagrees with the reference on some specific assertion), \emph{strictly stronger} (the candidate makes additional commitments), and \emph{strictly weaker} (the candidate drops a commitment).

\begin{table*}[t]
\centering
\setlength{\tabcolsep}{6pt}
\renewcommand{\arraystretch}{1.15}
\footnotesize
\begin{tabular}{lrlcccc}
\toprule
& & & \multicolumn{2}{c}{\textbf{Drop magnitude} [$95\%$ CI]} & \multicolumn{2}{c}{\textbf{Detection rate}} \\
\cmidrule(lr){4-5}\cmidrule(lr){6-7}
\textbf{Class} & $n$ & \textbf{Relation to ref.} & \textbf{SIV-recall} & \textbf{SIV-F1} & \textbf{recall} & \textbf{F1} \\
\midrule
\texttt{arg\_swap}                        & $494$ & incompatible      & $.818$ $[.796, .841]$ & $\mathbf{.757}$ $[.731, .784]$ & $.998$ & $\mathbf{.998}$ \\
\texttt{negation\_drop}                   & $132$ & incompatible      & $.861$ $[.820, .899]$ & $\mathbf{.820}$ $[.770, .868]$ & $1.000$ & $\mathbf{.992}$ \\
\texttt{random\_substitution}             & $638$ & incompatible      & $.960$ $[.949, .970]$ & $\mathbf{.942}$ $[.928, .956]$ & $1.000$ & $\mathbf{.998}$ \\
\texttt{flip\_outer\_quantifier}          & $355$ & strictly weaker   & $.984$ $[.975, .992]$ & $\mathbf{.975}$ $[.961, .986]$ & $1.000$ & $\mathbf{.997}$ \\
\texttt{restrictor\_drop}$^{\star}$       & $231$ & strictly stronger & $.000$ $[.000, .000]$ & $\mathbf{.169}$ $[.163, .175]$ & $.000^{\star}$ & $\mathbf{.996}$ \\
\texttt{strengthen\_quantifier}$^{\star}$ & $15$  & strictly stronger & $.000$ $[.000, .000]$ & $\mathbf{.325}$ $[.232, .439]$ & $.000^{\star}$ & $\mathbf{1.000}$ \\
\bottomrule
\end{tabular}
\caption{Six perturbation classes with per-class drop magnitude ($\text{score}(\text{ref}) - \text{score}(\text{perturbed})$, mean with $95\%$ bootstrap CIs) and within-pair detection rate. $\star$: strictly-stronger classes saturate SIV-recall at $1.0$ on both reference and perturbed candidates by construction, so SIV-F1's signal on these classes comes entirely from the contrastive arm.}
\label{tab:exp2-detection}
\end{table*}

\paragraph{Statistics.}
For each pair, we report \emph{drop magnitude} ($\text{score}(\text{ref}) - \text{score}(\text{perturbed})$, with $95\%$ bootstrap CIs over the class) and within-pair \emph{detection rate} (fraction of pairs in the class with $\text{score}(\text{ref}) > \text{score}(\text{perturbed})$).

\subsection{Results}

\textbf{SIV detects every class in over $99\%$ of pairs.} SIV-F1 mean drops range from $0.169$ on \texttt{restrictor\_drop} to $0.975$ on \texttt{flip\_outer\_quantifier}; all $95\%$ CIs exclude zero, and all detection rates exceed $0.99$ (Table~\ref{tab:exp2-detection}). Drop magnitudes calibrate to perturbation severity: large structural changes (a quantifier flip, a random predicate substitution) produce large drops; small structural changes (dropping a single restrictor conjunct) produce small drops.

\textbf{On strictly-stronger classes, the contrastive arm carries the entire detection signal.} A strictly-stronger candidate entails the reference, hence every positive probe by transitivity, so SIV-recall registers no drop at all on \texttt{restrictor\_drop} or \texttt{strengthen\_quantifier} (Table~\ref{tab:exp2-detection}, $\star$); their entire SIV-F1 drops ($0.169$ and $0.325$; detection $0.996$ and $1.000$) are sourced from contrastive probes the prover certifies as strictly stronger than $\varphi$. Removing the contrastive arm would drop detection on both classes to chance.

Two points connect back to \S\ref{sec:exp1-robust}: \texttt{random\_substitution} ($n{=}638$) lies entirely outside SIV's contrastive inventory yet is detected at $0.998$, and graded-LT detects every class here yet assigns \texttt{flip\_outer\_quantifier} and \texttt{restrictor\_drop} the same $0.5$, where SIV's drops separate them by a factor of six.

\section{Experiment 3: Error-class recovery from the trace}
\label{sec:exp3}

The previous two experiments measured the SIV score. This experiment evaluates what its labeled trace adds: whether the pattern of per-probe verdicts identifies which structural error occurred.

\subsection{Setup}

\paragraph{Classifier.}
A deterministic seven-rule classifier maps the (failed-positive-probes, fired-contrastive-probes) trace of each candidate to a perturbation class. Rules are applied in order; first match wins (Appendix~\ref{app:diagnostic-classifier}). Each rule implements a pre-registered prediction: each class declares its expected trace signature, and the classifier checks for that signature. There are no learned parameters, no fitted thresholds, no per-class hyperparameters.

\paragraph{Baseline.}
A $5$-fold stratified cross-validated decision tree on the single feature SIV-F1: the most the scalar score alone can extract about the perturbation class. The headline statistic is the recoverability \emph{delta} between trace classifier and score-only baseline.

\begin{table}[t]
\centering
\setlength{\tabcolsep}{3pt}
\renewcommand{\arraystretch}{1.15}
\footnotesize
\begin{tabular}{lrcccc}
\toprule
\textbf{Class} & $n$ & \textbf{F1} & \textbf{P} & \textbf{R} & \textbf{R}$_{\,\text{95\% CI}}$ \\
\midrule
\texttt{restrictor\_drop} & $231$ & $.996$ & $.996$ & $.996$ & $.976$--$1.00$ \\
\texttt{strengthen\_q}    & $15$  & $.933$ & $.933$ & $.933$ & $.681$--$.998$ \\
\texttt{negation\_drop}   & $132$ & $.778$ & $1.00$ & $.636$ & $.548$--$.718$ \\
\texttt{random\_sub}      & $638$ & $.634$ & $.464$ & $1.00$ & $.994$--$1.00$ \\
\texttt{arg\_swap}        & $494$ & $.482$ & $1.00$ & $.318$ & $.277$--$.361$ \\
\texttt{flip\_outer\_q}   & $355$ & $.006$ & $1.00$ & $.003$ & $.000$--$.016$ \\
\bottomrule
\end{tabular}
\caption{Per-class trace classifier performance. R$_{\,\text{95\% CI}}$: Clopper-Pearson interval. Analyses of \texttt{arg\_swap} and \texttt{flip\_outer\_q} in \S\ref{sec:exp3-stratum} and \S\ref{sec:exp3-flipquant}.}
\label{tab:exp3-perclass}
\end{table}

\paragraph{Pool.}
Experiment 2's $1{,}865$ verified pairs, scored by the full SIV pipeline of \S\ref{sec:metric}.

\subsection{Results}

\textbf{The trace nearly doubles diagnostic recovery over the score.} The rule classifier reaches macro-F1 $= 0.638$ at six-class granularity, against $0.333$ for the score-only baseline. This is a $+0.305$ recoverability delta; the unrecognized rate (Rule 7 fires) is $0.05\%$.

\textbf{Per-class breakdown.} Three classes are essentially solved (Table~\ref{tab:exp3-perclass}): \texttt{restrictor\_drop} at F1 $= 0.996$, \texttt{strengthen\_quantifier} at $0.933$, and \texttt{negation\_drop} at $0.778$ (perfect precision; some negation-drops fall to the catch-all rule). \texttt{random\_substitution}, the catch-all destination when no operator-specific signature fires, reaches $0.634$. Two classes need analysis: \texttt{arg\_swap} at $0.482$ (\S\ref{sec:exp3-stratum}) and \texttt{flip\_outer\_quantifier} at $0.006$ (\S\ref{sec:exp3-flipquant}).

\subsection{\texttt{arg\_swap}: ground atoms vs.\ universal bodies}
\label{sec:exp3-stratum}

\texttt{arg\_swap}'s aggregate F1 hides a clean bimodal pattern (Table~\ref{tab:exp3-argswap-stratum}): the trace recovers swaps reliably when the swapped relation occurs as a ground atom (stratum $S_5$, recall $0.882$, $n=127$) and fails when it is embedded in a universally-quantified body ($S_2$--$S_4$, recall $0.08$--$0.13$). The failure is a logical limit, not an implementation artifact: the asymmetry axiom $\forall x, y.\,R(x,y) \to \neg R(y,x)$ certifies a swap only when the formulas jointly assert $R(a,b) \wedge R(b,a)$ for concrete $a, b$; under a universal, both formulas are vacuously satisfiable, so Vampire correctly returns ``logically independent'' and the contrastive is never admitted into $\mathcal{P}^-$ (full account in Appendix~\ref{app:diagnostic-classifier}).

\subsection{\texttt{flip\_outer\_quantifier}: detected, not named}
\label{sec:exp3-flipquant}

\texttt{flip\_outer\_quantifier}'s F1 of $0.006$ is a coverage gap in the diagnostic layer, not in SIV's score, which detects these flips reliably (Experiment 1 AUC $= 1.000$; Experiment 2 mean drop $0.975$, detection $0.997$). The layer cannot \emph{name} the failure because the \texttt{flip\_quantifier} contrastive is admitted on only $0.3\%$ of these pairs, for the same vacuous-satisfaction reason as \S\ref{sec:exp3-stratum}. Closing the gap requires richer witness machinery (e.g., restrictor-existence axioms that force witnesses into both scopes); we defer it to future work to avoid post-hoc tuning to the experimental pool.

\section{Experiment 4: Real translator outputs}
\label{sec:exp4}

The controlled pools establish calibration; this experiment tests SIV on uncontrolled LLM translator outputs, measuring discrimination against expert correctness judgments, behavior on expert-labeled major errors, and behavior when the translator's vocabulary diverges from the reference.

\subsection{Setup}

We sample $150$ FOLIO premises (stratified across the \S\ref{sec:exp1} formula strata, seed $42$) and translate each with three configurations spanning the quality range: GPT-4.1 \citep{openai-2025-gpt41} free-vocabulary, GPT-4.1 with the reference predicate signature supplied (the high-accuracy regime of \citealp{brunello-2026-llms}), and Qwen2.5-7B \citep{qwen-2025-qwen25} free-vocabulary. Of $450$ outputs, $434$ parse. Labels are expert-audited: an independent LLM screening judge (a different model family from all translators) flags candidate errors; an expert labels $121$ pairs on blind sheets, including every flagged pair, so every ``incorrect'' label is human-verified; disputes were jointly adjudicated. Appendix~\ref{app:audit} gives the full protocol, agreement statistics, and per-configuration results.

\begin{table}[t]
\centering
\setlength{\tabcolsep}{6pt}
\renewcommand{\arraystretch}{1.15}
\footnotesize
\begin{tabular}{lcc}
\toprule
\textbf{Metric} & \textbf{AUC} & \textbf{$95\%$ CI} \\
\midrule
\rowcolor{blue!8}\textbf{SIV} & $\mathbf{0.841}$ & $[0.707, 0.935]$ \\
Smatch++  & $0.814$ & $[0.715, 0.899]$ \\
LE        & $0.799$ & $[0.650, 0.922]$ \\
LT        & $0.791$ & $[0.754, 0.826]$ \\
BLEU      & $0.708$ & $[0.533, 0.857]$ \\
BERTScore & $0.668$ & $[0.526, 0.796]$ \\
\bottomrule
\end{tabular}
\caption{AUC for separating expert-labeled correct from incorrect translations on the compatible-vocabulary stratum ($n = 185$, of which $10$ incorrect; expert-adjudicated labels).}
\label{tab:exp4-auc}
\end{table}

\subsection{Results}

\textbf{Discrimination.} SIV attains the top AUC point estimate ($0.841$; Table~\ref{tab:exp4-auc}), though with only $10$ incorrect pairs in this stratum the pairwise differences against Smatch++ and LE are not significant: on binary discrimination the prover- and graph-based metrics are at parity here. SIV's advantages are the two capabilities the AUC does not measure.

\textbf{SIV detects \emph{and grades} major errors.} Mean scores on expert-\textsc{correct} vs.\ expert-\textsc{major} pairs: SIV $0.80$ vs.\ $0.20$; LE $0.89$ vs.\ $0.62$; BERTScore $0.89$ vs.\ $0.82$ (barely moved); binary LT $0.58$ vs.\ $0.00$ (detects, but grades nothing). SIV is the only metric that does both.

\textbf{SIV knows its own scope.} $71$--$78\%$ of free-vocabulary translations use predicate vocabulary divergent from the reference; alignment (Appendix~\ref{app:alignment}) resolves $59$--$65\%$ of symbols, and SIV routes the genuinely out-of-vocabulary remainder to an abstention flag (F1 $0.87$, precision $0.93$) rather than silently mis-scoring it; no baseline has an analogue. The trade-off is explicit: divergent-but-correct pairs receive a low raw score ($0.29$) plus the flag, whereas BERTScore assigns such pairs $0.81$ and major errors $0.82$, ``handling'' divergence only by failing to distinguish it from error. With the signature supplied, $100\%$ of predicates align exactly.

\textbf{The inventory's boundary, located.} Of $29$ adjudicated errors, $14$ are vocabulary divergence (routed to abstention) and $15$ are in-scope. One is a strict over-strengthening the inventory does not generate: a conditional rendered as a biconditional, which saturates the positive arm, fires no contrastive, and scores $1.0$. A verified eleventh operator, \texttt{implication\_to\_biconditional}, closes the gap within the existing admission regime ($1.00 \to 0.889$ with a named trace label, no false fires on faithful conditionals; Appendix~\ref{app:operator-provenance}). A trace spot-check on ten flagged translations is in Appendix~\ref{app:audit}.

\textbf{FOLIO gold errors.} The audit independently surfaced $16/150$ premises (${\sim}11\%$) with erroneous gold formulas, corroborating \citet{olausson-etal-2023-linc}; excluding them, SIV's AUC rises to $0.855$ (still the top estimate). The $434$ audited pairs are released with the code.

\section{Conclusion}
\label{sec:conclusion}

SIV evaluates an NL$\rightarrow$FOL translation against two probe sets derived from the reference: positive probes the candidate must entail and contrastive probes it must not. The F1 over per-probe verdicts is the score; the labeled verdicts are the trace. On a stratified FOLIO pool, severity tier explains $80\%$ of SIV's variance against at most $17\%$ for any baseline, and the calibration survives merged tiers, inventory-disjoint operators, and a root-negation boundary probe; on a disjoint per-class pool, SIV detects every error class in over $99\%$ of pairs; the trace recovers the perturbation class at macro-F1 $0.638$ against $0.333$ score-only. On $434$ expert-audited real LLM translations, SIV attains the top AUC, uniquely detects \emph{and grades} major errors, and abstains on out-of-vocabulary pairs; the score ports to MALLS with zero FOLIO metadata (Appendix~\ref{app:portability}). A graded-LT ablation separates the credit: the prover supplies detection; gradation and diagnosis come from the probe decomposition.

Three extensions follow: syntactic-signature admission of \texttt{arg\_swap} contrastives (\S\ref{sec:exp3-stratum}); restrictor-existence axioms certifying outer-quantifier flips (\S\ref{sec:exp3-flipquant}); and the verified eleventh operator closing the audit's single in-scope gap (\S\ref{sec:exp4}, Appendix~\ref{app:operators}).

\clearpage
\section*{Limitations}

\paragraph{Scope of the claim.}
SIV is a \emph{structural-fidelity} metric for FOL translations within an alignable logical vocabulary; it is not a general-purpose metric for arbitrary NL$\rightarrow$FOL outputs, and we make no claim about arbitrary semantic equivalence. The metric's edit-operation taxonomy is the one used by \citet{brunello-2026-llms} and \citet{thatikonda-etal-2025-fol-metrics}, and the test-suite methodology follows \citet{zhong-etal-2020-semantic}; probe inventory and perturbation pools therefore share structural axes, as is standard for test-suite evaluation, though \S\ref{sec:exp1-robust} shows the calibration does not depend on that overlap. Our experiments demonstrate calibration under controlled perturbations and discrimination, grading, and abstention on real translator outputs (\S\ref{sec:exp4}); they do not show dominance on arbitrary candidate distributions. Errors outside the structural taxonomy, most prominently vocabulary divergence, are flagged by the abstention channel rather than scored (abstention F1 $0.87$); predicate-arity refactoring and alternative decomposition of compound concepts remain undetected, and a complete evaluation would pair SIV with vocabulary- or term-level components.

\paragraph{Operator coverage.}
The ten-operator contrastive inventory is not complete, and no perturbation taxonomy can be. The boundary is instead measured and extensible: on $434$ real translations, exactly one in-scope error class fell outside the inventory (conditional-to-biconditional over-strengthening, scored $1.0$), and a verified eleventh operator closes it within the existing admission regime (\S\ref{sec:exp4}, Appendix~\ref{app:operators}). Users applying SIV outside its validated scope should treat high scores on structurally exotic candidates with care; the positive arm still catches any candidate that loses or contradicts reference content, but overstatement detection is only as broad as the admitted contrastive set.

\paragraph{Equal probe weighting and human severity judgments.}
All probes carry equal weight in the F1. This is a deliberate first choice: any severity weighting embeds a downstream-task judgment we did not want to hard-code into a reference-based metric. The F1 already carries an implicit structural weighting, since an error in central content fails the reference-itself probe plus every probe downstream of the damaged node, which is why Experiment 2's drop magnitudes track severity ($0.169$ to $0.975$); and because probes are labeled, a task-specific weighted SIV is a drop-in variant (weights over probe labels), which we leave to future work. Relatedly, our evidence for correspondence with \emph{fine-grained} human severity judgments is limited: the expert audit validates the binary half (all \S\ref{sec:exp4} statistics are measured against human correctness labels), but real errors were overwhelmingly major ($2$ in-scope minor errors in $434$ pairs after adjudication), so a graded correlation against expert severity is underpowered on real data. The controlled Experiment-1 calibration remains the graded evidence, anchored at both ends by real-output behavior.

\paragraph{Auxiliary axioms and dataset-specific metadata.}
Probe admission runs Vampire with default axioms together with per-predicate and per-restrictor existence witnesses; for binary relations, asymmetry and symmetry axioms come from a frozen $427$-predicate label table built from FOLIO's vocabulary ($850$ of $877$ observed predicate usages covered). The dependence is measured (Appendix~\ref{app:portability}): the table enters SIV at exactly one code path (\texttt{swap\_binary\_args} admission); disabling it leaves per-class \emph{detection} unchanged to three decimals in all six classes, and on MALLS, with zero FOLIO metadata, the parser accepts $96.6\%$ of golds and per-class detection is $1.000/0.969/1.000$. What the table buys is diagnostic \emph{naming}: without it, arg-swap trace-F1 falls from $0.482$ to $0.000$ and macro-F1 from $0.638$ to $0.550$ (still $+0.189$ over the score-only baseline). Porting SIV's score requires no metadata; only the trace's arg-swap axis needs a domain label table or the \S\ref{sec:exp3-flipquant} witness extension.

\paragraph{FOLIO reference errors.}
FOLIO's expert annotations contain approximately $11\%$ errors \citep{olausson-etal-2023-linc, brunello-2026-llms}; our real-output audit independently corroborated this rate ($16/150$ premises, Appendix~\ref{app:audit}), and excluding those premises raises SIV's AUC to $0.855$. On a prior pool, manual correction of the top-30 most-suspect references shifted headline statistics by $\pm 0.02$ (Appendix~\ref{sec:reference-corrections}); we have not re-audited the present stratified pool and treat the prior stability as indicative.

\paragraph{Pool composition.}
Pool composition in Experiment 1 concentrates some tiers on a small set of operators (\textsc{OS}: $69\%$ from \texttt{OS\_add\_nucleus\_conjunct}; \textsc{P}: $n=25$, $72\%$ from \texttt{P\_drop\_conjunct}). Per-operator AUC (Appendix~\ref{app:per-operator-auc}) shows the metric ordering is consistent across operators, so the headline statistics are not driven by any single operator. In Experiment 2, the \texttt{strengthen\_quantifier} class yields $15$ verified pairs after Vampire filtering; the architectural-payoff finding rests primarily on \texttt{restrictor\_drop} ($n=231$).

\paragraph{Diagnostic admissibility ceiling.}
The diagnostic classifier names a structural failure only when the corresponding contrastive is admitted into $\mathcal{P}^-$. Under our admissibility regime, \texttt{flip\_quantifier} contrastives are admitted on $0.3\%$ of outer-quantifier-flip pairs and \texttt{swap\_binary\_args} on $\sim 13\%$ of universal-body swaps. Per-class diagnostic performance is correspondingly uneven (Table~\ref{tab:exp3-perclass}): \texttt{restrictor\_drop} at F1 $0.996$ and \texttt{strengthen\_quantifier} at $0.933$, against \texttt{flip\_outer\_quantifier} at $0.006$. The score detects both perturbation types reliably (Experiments 1 and 2); only the diagnostic layer's per-class naming is affected. Closing these gaps requires extending the witness machinery (\S\ref{sec:exp3-flipquant}).

\bibliography{custom}

\appendix
 
\section{Severity Tier Construction}
\label{app:tier-construction}
 
The Experiment~1 candidate pool is constructed via a stratified, pre-registered design. The design predates generation and was not revised against observed metric outcomes; the configuration is included in the reproduction package (Appendix~\ref{app:reproducibility}).
 
\paragraph{Tier definitions.}
The four severity tiers are defined in Table~\ref{tab:tiers} of the main text. Tier labels are determined by Vampire-verified logical relation plus a structural skeleton check, not by LLM self-assessment. The \textsc{P}/\textsc{OW} boundary is operationalized as a structural AST property: walking the candidate and reference ASTs, we compare the top-level quantifier and connective sequence (ignoring predicate names, argument lists, and inner-clause content). A candidate that preserves this sequence is classified \textsc{P}; otherwise \textsc{OW}.
 
\paragraph{Generation.}
Candidates are generated by applying $14$ typed perturbation operators across the five formula strata $\{S_1, \ldots, S_5\}$. Each operator declares its target tier (\textsc{OS}/\textsc{P}/\textsc{OW}) and an applicability predicate; the design specifies a per-(operator $\times$ stratum) cell target. Operators include \textsc{OS}-tier strengthening operators (\texttt{OS\_add\_nucleus\_conjunct}, \texttt{OS\_narrow\_consequent}, \texttt{OS\_strengthen\_predicate}, \texttt{OS\_strengthen\_quantifier}, \texttt{OS\_drop\_conjunctive\_restrictor}), \textsc{P}-tier weakening operators (\texttt{P\_drop\_conjunct}, \texttt{P\_weaken\_predicate}, \texttt{P\_drop\_disjunctive\_restrictor}), and \textsc{OW}-tier scaffolding-damage operators (\texttt{OW\_flip\_outer\_quantifier}, \texttt{OW\_de\_quantify\_to\_c0}, \texttt{OW\_weaken\_to\_existential}, \texttt{OW\_overrestrict\_antecedent}, \texttt{OW\_drop\_consequent\_severely}, \texttt{OW\_weaken\_predicate\_severely}).
 
\paragraph{Verification.}
Each generated candidate is verified by Vampire in both directions ($\textsc{cand} \models \textsc{ref}$ and $\textsc{ref} \models \textsc{cand}$) and, where applicable, by the structural skeleton-match check. A candidate is retained iff its verdict matches its declared tier. Verification retention: $547$ generated, $529$ retained ($96.7\%$).
 
\paragraph{Cell-target outcomes.}
Several (operator $\times$ stratum) cells have target $8$ but $0$ retained, because Vampire dropped all generated candidates when the operator's applicability predicate did not fire on the stratum (e.g., \texttt{OS\_strengthen\_predicate} on $S_4$ nested-quantifier; \texttt{P\_weaken\_predicate} on $S_2$ universal-simple). These outcomes are disclosed pool composition.
 
\paragraph{Pre-registration discipline.}
Three $S_2$-cell operators (\texttt{OW\_weaken\_to\_existential}, \texttt{OW\_overrestrict\_antecedent}, \texttt{P\_drop\_disjunctive\_restrictor}) were initially excluded by a bookkeeping error; their applicability predicates do fire on $\forall$-implications with non-atomic consequents. The $56$ affected candidates were correctly generated and Vampire-verified before this was caught, and no scoring had occurred when they were added back. We disclose this because pre-registration discipline depends on noting late changes to the design.

\paragraph{The absent stronger-and-skeleton-breaking cell.}
The design space contains a cell the tier system does not instantiate: perturbations that are strictly \emph{stronger} and also break the top-level skeleton. The tiers order structural distance for single typed edits, the errors translators actually make, and a strengthening edit that also breaks the top-level skeleton requires compound edits; the cell is correspondingly rare as a single-edit translation error, which is why the pool lacks it. The root-negation boundary probe of \S\ref{sec:exp1-robust} tests the adjacent extreme (whole-formula complements) directly.
 
\section{Per-Operator AUC Decomposition}
\label{app:per-operator-auc}
 
We report per-operator AUC for separating \textsc{Ref} from each operator's candidate set, per metric. Two operators localize the LE propositional collapse documented in \S\ref{sec:exp1}:
 
\begin{table}[h]
\centering
\setlength{\tabcolsep}{3pt}
\renewcommand{\arraystretch}{1.15}
\footnotesize
\begin{tabular}{lcc}
\toprule
\textbf{Operator} & \textbf{LE AUC} & \textbf{SIV AUC} \\
\midrule
\texttt{OW\_flip\_outer\_quantifier} & $0.500$ & $\mathbf{1.000}$ \\
\texttt{OS\_strengthen\_quantifier}  & $0.500$ & $\mathbf{1.000}$ \\
\bottomrule
\end{tabular}
\caption{LE is at chance on both quantifier-manipulation operators; SIV saturates at $1.000$. This reproduces \citet{brunello-2026-llms}'s propositional-collapse prediction operator by operator on the stratified pool.}
\label{tab:per-operator-le-failure}
\end{table}
 
LE's AUC values on the remaining operators range from $0.66$ to $1.00$. The chance behavior is specific to operators that manipulate quantifier scope or type without altering predicate vocabulary: LE's truth-table representation cannot detect changes that do not register at the propositional level.

\section{Robustness Checks for Experiment 1}
\label{app:robustness}

This appendix details the four checks summarized in \S\ref{sec:exp1-robust}.

\paragraph{Merging the equal-rank tiers.} \textsc{OS} and \textsc{P} share severity rank by design (\S\ref{sec:exp1}), so one may ask whether the headline $\eta^2$ depends on splitting tiers the design calls equal. Recomputing $\eta^2$ with \textsc{OS} and \textsc{P} merged into a single level, giving the statistic zero credit for separating them, yields SIV $\eta^2 = 0.706$, against at most $0.07$ for every baseline, a roughly $10\times$ margin that survives the merge entirely. The score asymmetry between \textsc{OS} and \textsc{P} is itself informative rather than inconsistent: an \textsc{OS} candidate still asserts everything the reference asserts (it errs by adding), while a \textsc{P} candidate fails to assert the reference's content (it errs by losing), and a recall-bearing F1 prices lost content more heavily. At the per-premise level, SIV produces zero severity-order reversals across all $372$ candidates: it never ranks a more-broken tier above a less-broken one, and where it does not strictly separate, it ties.

\paragraph{Operators disjoint from SIV's inventory.} Only $3$ of SIV's $10$ contrastive operators are exercised by Experiment 1's $14$ pool operators at all; $9$--$10$ of the $14$ are structurally disjoint from the inventory (conjunct addition, predicate-hierarchy substitution, quantifier elimination, and others on no contrastive axis). Restricting Experiment 1 to the disjoint operators leaves SIV at $\eta^2 = 0.759$ (vs.\ $0.802$ on the full pool) and Cohen's $d = 4.37$ $[3.76, 5.49]$, against at most $0.16$ for every baseline, with non-overlapping CIs and AUC $= 1.00$ on every disjoint severe-tier operator. Experiment 2's \texttt{random\_substitution} class ($n = 638$) is likewise out-of-inventory and detected at $0.998$. SIV's validation was already mostly outside its own taxonomy.

\begin{table}[t]
\centering
\setlength{\tabcolsep}{3.4pt}
\renewcommand{\arraystretch}{1.15}
\footnotesize
\begin{tabular}{lcccccc}
\toprule
\textbf{Candidate} & \textbf{SIV} & \textbf{LE} & \textbf{LT} & \textbf{Sm++} & \textbf{BLEU} & \textbf{BS} \\
\midrule
root negation & $\mathbf{.022}$ & $.000$ & $.000$ & $.864$ & $.419$ & $.858$ \\
(\textsc{OW} mean) & $.052$ & $.713$ & $.000$ & $.740$ & $.397$ & $.822$ \\
\bottomrule
\end{tabular}
\caption{Root-negation boundary probe: the direct negation of the reference, scored for all $105$ Experiment-1 premises (Vampire-verified incompatible, $105/105$), with the \textsc{OW}-tier means for comparison. Sm++: Smatch++; BS: BERTScore.}
\label{tab:root-negation}
\end{table}

\paragraph{A root-negation boundary probe.} The tiers order structural distance for single typed edits, the errors translators actually make, so whole-formula complements were not a tier; the direct negation of the reference is the natural boundary probe. We score it for every Experiment-1 premise (Table~\ref{tab:root-negation}). SIV places the negation \emph{below} its lowest defined tier ($0.022$; the bootstrap CI's upper bound, $0.048$, is itself below the \textsc{OW} mean of $0.052$), so the hierarchy's endpoints extrapolate in the right direction without having been designed for this case. LE and LT also correctly score the negation at $0$: a root negation is exactly what a truth-table representation catches; this row is not evidence against them. The surface and graph metrics do not: Smatch++ scores the reference's logical \emph{complement} above every error tier, including over-strengthened but content-preserving translations.

\paragraph{Graded-LT construction.} The graded LT of \S\ref{sec:exp1-robust} is the strongest we could build: bidirectional entailment under SIV's exact prover setup (Vampire, same axioms and timeout), scored $1.0 / 0.5 / 0.0$ for equivalent / one-directional / neither, run on both experimental pools. It detects every Experiment-2 perturbation class in $100\%$ of pairs: the prover's strength is real. But it is a three-valued function of the logical \emph{relation}, so it scores the mildest error tier (\textsc{OS}) and the most severe, skeleton-broken tier (\textsc{OW}) identically at $0.5$; on Experiment 2 its only two drop values track relation type rather than severity, assigning \texttt{flip\_outer\_quantifier} (SIV's largest drop, $0.975$) the same $0.5$ as \texttt{restrictor\_drop} (SIV's smallest, $0.169$); and it produces no trace.

\section{Contrastive Operator Inventory}
\label{app:operators}
 
The contrastive probe set $\mathcal{P}^{-}(\varphi)$ is generated by applying ten operators at all valid sites of the reference. Each operator produces either a strictly-stronger contrastive ($\psi \models \varphi$ but $\varphi \not\models \psi$, used to detect under-constraining translations) or an incompatible contrastive ($\varphi \wedge \psi \models \bot$, used to detect drift along a specific structural axis).
 
\begin{itemize}
\setlength{\itemsep}{1pt}
\item \texttt{negate\_atom}: flips negation on a single atomic predicate. Incompatible.
\item \texttt{swap\_binary\_args}: reverses argument order in binary predicates ($R(a,b) \to R(b,a)$). Incompatible; admissibility requires asymmetry-axiom certification (\S\ref{sec:metric-contrastive}).
\item \texttt{flip\_quantifier}: swaps $\forall \leftrightarrow \exists$ on a single binder. Strictly-stronger when changing $\exists$ to $\forall$ inside an implication; incompatible otherwise. Admissibility on outer-quantifier flips is the disclosed coverage limit of \S\ref{sec:exp3-flipquant}.
\item \texttt{drop\_restrictor\_conjunct}: removes a single conjunct from a restrictor. Strictly-stronger.
\item \texttt{flip\_connective}: swaps $\wedge \leftrightarrow \vee$ at a single junction. Incompatible.
\item \texttt{replace\_subformula\_\allowbreak with\_negation}: wraps a subformula in negation. Incompatible.
\item \texttt{disjunct\_drop}: removes a disjunct from a disjunction. Strictly-stronger.
\item \texttt{converse}: reverses an implication's direction ($\alpha \to \beta \mapsto \beta \to \alpha$). Incompatible.
\item \texttt{scope\_swap}: permutes adjacent quantifiers of opposite type. Strictly-stronger.
\item \texttt{equality\_drop}: removes an equality literal from a conjunction. Strictly-stronger.
\end{itemize}
 
Each operator validates its output with a type-soundness check; ill-formed candidates are discarded. The full FOLIO suite contains $1{,}678$ strictly-stronger and $2{,}254$ incompatible generated contrastives across $1{,}393$ premises (mean $2.82$ per premise). The fraction \emph{admitted} into $\mathcal{P}^-$ depends on the active axiom set; admissibility rates per class and per operator are reported with the diagnostic experiment (\S\ref{sec:exp3}).
 
\paragraph{Asymmetry-axiom label table.}
The frozen $427$-predicate asymmetry/symmetry/unknown label table used to admit \texttt{swap\_binary\_args} contrastives is released alongside the code. The table covers $850$ of $877$ predicate usages observed in the experimental pools; the remaining $27$ usages map to the $21$ \texttt{unknown}-labeled predicates and contribute no axiom.

\subsection{Provenance, selection principles, and extensibility}
\label{app:operator-provenance}

\begin{table}[t]
\centering
\setlength{\tabcolsep}{3pt}
\renewcommand{\arraystretch}{1.2}
\scriptsize
\begin{tabular}{lll}
\toprule
\textbf{Operator} & \textbf{Axis} & \textbf{Provenance} \\
\midrule
\texttt{negate\_atom} & polarity (atom) & prior$^{a,b}$ \\
\texttt{flip\_quantifier} & quantifier type & prior$^{a,b}$ \\
\texttt{flip\_connective} & connective & prior$^{a,b}$ \\
\texttt{replace\_subf.\_w\_neg.} & polarity (subf.) & adapted$^{c}$ \\
\texttt{swap\_binary\_args} & argument order & new \\
\texttt{drop\_restrictor\_conj.} & restrictor & new$^{d}$ \\
\texttt{disjunct\_drop} & disjunction & new \\
\texttt{converse} & implication dir. & new \\
\texttt{scope\_swap} & quantifier scope & new \\
\texttt{equality\_drop} & equality & new \\
\midrule
\texttt{implic.\_to\_bicond.} & implication str. & new$^{e}$ \\
\bottomrule
\end{tabular}
\caption{Contrastive-operator provenance. $^{a}$\citet{brunello-2026-llms}; $^{b}$\citet{thatikonda-etal-2025-fol-metrics}; $^{c}$literal-level negation generalized to subformulas; $^{d}$derived from the Barwise--Cooper decomposition \citep{barwise-cooper-1981}, not from an error taxonomy; $^{e}$post-audit extension operator, verified but not used in the \S\ref{sec:exp1}--\S\ref{sec:exp4} experiments (see below).}
\label{tab:operator-provenance}
\end{table}

Table~\ref{tab:operator-provenance} records, for each operator, the structural axis it targets and whether it comes directly from a prior error inventory, is adapted from one, or is newly introduced here.

\paragraph{Selection principle.} The inventory was selected for systematic coverage of FOL's structural axes (quantifier type, scope, polarity, argument order, connective, restrictor, equality, implication direction), one operator per axis and two where granularities differ. The operators were \emph{not} curated from observed errors: none was added or shaped in response to errors seen during development. Each operator is only a generator; its declared relation to the reference must be certified by Vampire before admission (\S\ref{sec:metric-contrastive}), so the inventory is a closed, axis-motivated, prover-filtered set.

\paragraph{Measured boundary and a verified extension.} The real-output audit (\S\ref{sec:exp4}) located the inventory's boundary concretely: of $15$ in-scope structural errors, $1$ was a strict over-strengthening no current operator generates, a conditional rendered as a biconditional ($\forall x.(R(x) \to N(x))$ translated with $\leftrightarrow$), which saturates the positive arm, fires no contrastive, and receives SIV $1.0$. The fix stays inside the architecture, and we verified it: an eleventh operator, \texttt{implication\_to\_biconditional}, is admitted as strictly-stronger by Vampire under the existing witness regime, fires on the audited candidate (SIV falls from $1.00$ to $0.889$ with the trace naming the axis), and does not fire on the reference or on faithful conditional translations. We report it as a verified extension rather than folding it into the experimental inventory, so that every number in \S\ref{sec:exp1}--\S\ref{sec:exp4} reflects the pre-registered ten-operator set. The taxonomy is not complete; its boundary is measurable and extensible.
 
\section{Predicate Alignment Procedure}
\label{app:alignment}
 
Candidate translations may use predicate vocabulary that differs from the reference (e.g., \texttt{HasWheels} vs.\ \texttt{HasWheel}). We perform deterministic alignment via a similarity threshold of $0.6$, applied to embedding-based symbol matching with a layered fallback over exact match, lemma match, and substring match. Aligned candidate predicates are rewritten to the reference vocabulary before scoring; unaligned predicates are adjudicated against the candidate's original symbol.
 
\paragraph{Sensitivity.}
On the operator-driven Experiment~1 pool used for the headline statistics, alignment is effectively the identity map because candidates inherit canonical predicate names from the SIV inventory by construction. The threshold of $0.6$ was set during initial development and was not tuned against evaluation outcomes.

\paragraph{Behavior on real translator outputs.}
On the \S\ref{sec:exp4} pool, alignment behavior splits cleanly by translator configuration. With the reference signature supplied, $100\%$ of candidate predicates align exactly. Free-vocabulary translators diverge heavily ($71$--$78\%$ of pairs contain divergent vocabulary), and the layered alignment resolves $59$--$65\%$ of divergent symbols, leaving a remainder that is genuinely out of vocabulary.

\paragraph{Out-of-vocabulary abstention.}
When alignment leaves reference-critical predicates unresolved, SIV emits an out-of-scope \emph{abstention flag} for the pair instead of a confident score. On the \S\ref{sec:exp4} pool the flag identifies vocabulary-divergent pairs at F1 $0.87$ (precision $0.93$). This is deliberate: on divergent-but-correct pairs SIV's raw score is low ($0.29$), and the flag marks the score as ungrounded rather than pretending to adjudicate vocabulary the prover cannot align.
 
\section{Reference Correction Sensitivity}
\label{sec:reference-corrections}
 
We surface candidate reference errors via a weighted score combining four signals: SIV-soft recall $< 1.0$ on the reference (weight $100\times$), SIV-strict recall $< 1.0$ on the reference (weight $10\times$), predicate-Jaccard between SIV canonical and FOLIO gold $< 0.8$ (weight $5\times$), and SIV-soft F1 $< 1.0$ on the reference (weight $1\times$). On the prior submission's pool of $368$ premises, $69$ triggered at least one signal; the top $30$ by weighted score were inspected manually. Five confirmed transcription errors (entity mismatches between NL and gold formula, missing negations, reversed implications) were corrected and one premise with a typo in the source NL was excluded. Recomputing the prior submission's headline statistic on the corrected subset and on the full corpus with corrections applied yielded a stability of $\pm 0.02$. We have not re-run this correction pass on the new stratified Experiment~1 pool; the prior result is indicative of the FOLIO reference-error magnitude but not a direct sensitivity check for the new design. The \S\ref{sec:exp4} audit independently corroborates the magnitude: $16$ of $150$ stratified premises (${\sim}11\%$) carried erroneous gold formulas, matching the rate reported by \citet{olausson-etal-2023-linc}, and excluding them raises SIV's real-output AUC from $0.841$ to $0.855$.
 
\section{Diagnostic Classifier Rules}
\label{app:diagnostic-classifier}

\begin{table}[t]
\centering
\footnotesize
\setlength{\tabcolsep}{4pt}
\begin{tabular}{lrr}
\toprule
\textbf{Reference stratum} & $n$ & \textbf{Recall} \\
\midrule
$\mathbf{S_5}$\,~ground-atom relations & $127$ & $\mathbf{0.882}$ \\
$S_8$\,~other                          & $9$   & $0.667$ \\
$S_6$\,~negation                       & $16$  & $0.250$ \\
$S_3$\,~universal-multi-restrictor     & $176$ & $0.125$ \\
$S_2$\,~universal-simple               & $49$  & $0.082$ \\
$S_4$\,~nested-quantifier              & $86$  & $0.081$ \\
$S_7$\,~existential                    & $31$  & $0.065$ \\
\bottomrule
\end{tabular}
\caption{\texttt{arg\_swap} recovery by reference stratum (discussed in \S\ref{sec:exp3-stratum}).}
\label{tab:exp3-argswap-stratum}
\end{table}

\paragraph{Why universal bodies defeat the asymmetry route (\S\ref{sec:exp3-stratum}).}
Ground-atom occurrences trigger the axiom directly: the reference asserts $R(a,b)$, the swap asserts $R(b,a)$, asymmetry forbids the conjunction, and Vampire admits the swap as incompatible. In a universally-quantified body, neither formula asserts a specific $R$-fact: both are conditional and vacuously satisfied in any model where the antecedent is empty, so asymmetry never fires and the classifier loses its signal. This is a limit of asymmetry axioms under standard FOL semantics, not a Vampire timeout or a missing predicate-label entry.

The diagnostic classifier (\S\ref{sec:exp3}) maps the labeled trace $(\text{failed\_positives}, \text{fired\_contrastives})$ to a perturbation class via the following rules, applied in order; first match wins.
 
\begin{enumerate}
\setlength{\itemsep}{1pt}
\item \texttt{drop\_restrictor\_conjunct} fired AND failed\_positives empty $\to$ \texttt{restrictor\_drop}.
\item \texttt{flip\_quantifier} fired AND failed\_positives empty $\to$ \texttt{strengthen\_quantifier}.
\item \texttt{flip\_quantifier} fired AND failed\_positives non-empty $\to$ \texttt{flip\_outer\_quantifier}.
\item \texttt{swap\_binary\_args} fired $\to$ \texttt{arg\_swap}.
\item \texttt{negate\_atom} OR \texttt{replace\_subformula\_\allowbreak with\_negation} fired $\to$ \texttt{negation\_drop}.
\item failed\_positives non-empty AND no contrastives fired $\to$ \texttt{random\_substitution}.
\item Otherwise $\to$ \texttt{unrecognized}.
\end{enumerate}
 
The rule table implements a pre-registered prediction pattern per class (declared in advance of any scoring): each class names its expected (failed\_positives, fired\_contrastives) signature, and the classifier checks for that signature. There are no learned parameters, fitted thresholds, or per-class hyperparameters.
 
\section{Detailed Probe Statistics}
\label{sec:probe-stats}
 
The full FOLIO evaluation suite contains $1{,}393$ of $1{,}471$ premises with at least one well-formed probe ($94.7\%$); $2{,}340$ positive probes (mean $1.68$ per premise); and $3{,}932$ generated contrastive probes (mean $2.82$ per premise; $2{,}254$ incompatible, $1{,}678$ strictly-stronger). Total: $6{,}272$ structural probes generated. The admitted-into-$\mathcal{P}^-$ count depends on the active axiom set (\S\ref{sec:metric-contrastive}, \S\ref{sec:exp3}). Full suite construction includes additional audit calls reaching $7{,}934$ Vampire invocations, with $0$ hard failures across the audit. Of the $156$ references containing a disjunction, none contributes a positive probe by disjunct removal; disjunctions enter $\mathcal{P}^{+}$ only carried whole (\S\ref{sec:metric-positive}).
 
\section{Real-Translator Audit: Protocol and Detailed Results}
\label{app:audit}

\paragraph{Translator configurations.}
The three configurations of \S\ref{sec:exp4} span the quality range documented by \citet{brunello-2026-llms}: GPT-4.1 with free predicate vocabulary, GPT-4.1 with the reference predicate signature supplied in the prompt (the high-accuracy regime of that study), and Qwen2.5-7B with free vocabulary. Each translates the same $150$ stratified FOLIO premises (seed $42$). Of $450$ outputs, $434$ parse; the malformed counts are $3$, $0$, and $13$ respectively.

\paragraph{Labeling protocol.}
An independent LLM screening judge, drawn from a different model family than all three translators, first flags candidate errors. An expert annotator then labels $121$ pairs directly on blind sheets (translator identity and metric scores hidden), including \emph{every} pair the judge flagged as incorrect, so the entire ``incorrect'' class in the analysis is human-verified. A second human annotator double-annotated a $50$-pair overlap: human--human agreement on the binary correct/error boundary is $\kappa = 0.68$ ($84\%$ raw agreement), identical to expert--judge agreement ($\kappa = 0.68$, $87\%$ raw). Agreement on the \textsc{major}/\textsc{minor} severity boundary is lower ($\kappa = 0.57$), so every pair with a disputed label was jointly adjudicated before analysis.

\paragraph{Error composition.}
After adjudication, the audit contains $29$ confirmed errors: $14$ vocabulary-divergence errors, which SIV routes to the abstention flag (Appendix~\ref{app:alignment}), and $15$ in-scope structural errors, of which $2$ are minor and $1$ is the conditional-to-biconditional over-strengthening outside the ten-operator inventory (Appendix~\ref{app:operator-provenance}). The scarcity of in-scope \emph{minor} errors ($2$ in $434$ pairs) is why a graded correlation against expert severity is underpowered on real data (Limitations).

\paragraph{Compatible-vocabulary stratum.}
The AUC comparison of Table~\ref{tab:exp4-auc} is computed on the stratum where all metrics are defined without vocabulary confounds: $n = 185$ pairs with compatible vocabulary, of which $10$ are expert-labeled incorrect. With $10$ positives, pairwise AUC differences against Smatch++ and LE are not significant; SIV's mean-score separation on expert-\textsc{major} vs.\ expert-\textsc{correct} pairs ($0.20$ vs.\ $0.80$) and its abstention flag are the capabilities the AUC comparison does not capture.

\paragraph{Trace spot-check.}
We spot-checked the diagnostic trace on $10$ flagged translations. Two proved correct on adjudication; one of these was a confirmed FOLIO gold error, on which SIV had scored $1.00$ and stayed silent, consistent with that label. Of the $8$ confirmed errors, the trace named the exact structural axis in $1$, localized the broken content through failed positive probes in $6$, and was silent in $1$, the biconditional coverage gap now closed by the verified eleventh operator. The axis-naming case is instructive: for the FOLIO premise ``In Love City, everyone is scared of animals, or loves animals, or both,'' GPT-4.1 dropped the antecedent restricting the rule to Love City, turning it into a claim about \emph{everyone}, and the trace fired a single contrastive, \texttt{drop\_restrictor\_conjunct}, naming exactly that error.

\paragraph{Released artifacts.}
The $434$ audited (premise, translation, label) triples, the blind annotation sheets, and the per-pair metric scores are released with the code.

\section{Metadata Ablation and MALLS Portability}
\label{app:portability}

We tested the dependence on FOLIO-specific metadata from two directions: in-domain, by disabling the $427$-predicate table on our own pools and measuring what breaks, and cross-domain, by running SIV on MALLS \citep{yang-etal-2024-mallspaper}, where the table barely applies. The two agree.

\paragraph{In-domain ablation (FOLIO, table disabled).}
The $427$-predicate table enters SIV at exactly one code path: \texttt{swap\_binary\_args} admission. Removing it deletes $336$ of $2{,}553$ admitted contrastives, all on that one operator, and per-class \emph{detection} is unchanged to three decimals in all six \S\ref{sec:exp2} classes (\texttt{arg\_swap} $0.998$ with or without the table; SIV-recall exactly invariant across all $2{,}505$ scored candidates). What the table buys is diagnostic \emph{naming}: arg-swap trace-F1 falls from $0.482$ to $0.000$ (misrouting to the catch-all class), and macro-F1 from $0.638$ to $0.550$, though the trace still beats the score-only baseline by $+0.189$ with zero metadata.

\paragraph{Cross-domain probe (MALLS, zero FOLIO metadata).}
With no domain adaptation of any kind, the parser accepts $96.6\%$ of MALLS gold formulas. Probe density is \emph{higher} than on FOLIO ($5.65$ vs.\ $2.82$ admitted contrastives per premise), so the mechanism does not thin out without metadata. Per-class detection on the three ported classes is $1.000$ / $0.969$ / $1.000$ (negation-drop / restrictor-drop / outer-quantifier-flip), with restrictor-drop reproducing the architectural-payoff finding of \S\ref{sec:exp2} on a new domain: recall saturates and the contrastive arm alone carries detection. Enabling the full FOLIO table on MALLS changes admission by $2$ probes out of $565$; it covers $8\%$ of MALLS's binary predicates and is genuinely FOLIO-specific, but nothing that carries detection depends on it.

\paragraph{Summary.}
Porting SIV's \emph{score} requires no metadata at all; only the trace's arg-swap axis needs a domain label table or the witness extension of \S\ref{sec:exp3-flipquant}.

\section{Runtime Profile}
\label{app:runtime}

\paragraph{Decomposition.}
Profiling on the \S\ref{sec:exp2} pool (14-core workstation): probe construction plus Vampire admission is paid once per reference ($\sim 0.35$\,s, ${\sim}24$ prover calls) and reused across every candidate and every system evaluated. The marginal cost of scoring one additional candidate is $67$\,ms single-core and $6.8$\,ms parallel (${\sim}5.7$ prover calls, median $12$\,ms, zero timeouts in $14{,}293$ calls). Evaluating $N$ systems on a benchmark costs build-once plus $N \times$ seconds, which is how the released pipeline already works: suites are prebuilt and loaded.

\paragraph{Cheaper variants, and a pitfall.}
For iterative loops we profiled two cost reducers. Reducing the per-call timeout is inert: no call approaches it. A naive one-probe-per-axis cap buys $1.8\times$ but silently breaks the \texttt{restrictor\_drop} detection class ($0.996 \to 0.017$), because it discards all but one probe of each restrictor group; a corrected cap that keeps restrictor groups intact restores detection above $0.99$ at about $1.6\times$.

\section{Reproducibility}
\label{app:reproducibility}
 
\paragraph{Theorem prover.}
All probe adjudication uses Vampire v$5.0.1$ with a per-call timeout of $5$ seconds. The release script auto-downloads the pinned binary; CI verifies binary checksum. Timeouts are treated as ``not entailed'' for both positive and contrastive probes in the strict variant, and symmetrically in the soft variant; we report the soft variant in the main text (\S\ref{sec:metric}).
 
\paragraph{Reproduction snapshot.}
{\sloppy All experiments are reproduced from the single code release linked in \S\ref{sec:intro}. Each is a one-command invocation against its design artifact: \texttt{configs/\allowbreak severity\_\allowbreak correlation\_v1.yaml} for Experiment~1, \texttt{run\_perturbation\_detection.py} for Experiment~2, \texttt{run\_diagnostic\_trace.py} for Experiment~3, and \texttt{run\_real\_\allowbreak translator\_\allowbreak audit.py} for Experiment~4, alongside orchestrators for the robustness checks of \S\ref{sec:exp1-robust} and the ablations of Appendices~\ref{app:portability} and~\ref{app:runtime}. The 427-predicate asymmetry/symmetry label table used by the diagnostic experiment is shipped with the code as \texttt{siv/asymmetry\_axioms.py}; the $434$ expert-audited real-translation pairs ship under \texttt{data/audit/}. Per-experiment artifacts (per-tier means, drop magnitudes, classifier predictions, contrastive-firing matrix) are regenerated by re-running the orchestrators; reference outputs are included under \texttt{reports/experiments/} for direct comparison.\par}
 
\paragraph{Random seeds.}
All deterministic experiments are seed-independent given fixed Vampire output. Pool construction uses seed $42$ for within-cell ordering and for any tie-breaking; bootstrap CIs use $1{,}000$ resamples with seed $42$. The diagnostic classifier (\S\ref{sec:exp3}) is rule-based and has no random component; the score-only baseline uses 5-fold stratified cross-validation with seed $42$.
 
\paragraph{Alignment threshold.}
The predicate-alignment threshold (Appendix~\ref{app:alignment}) is fixed at $0.6$ across all experiments and was not tuned against evaluation outcomes.
 
\end{document}